\documentclass[11pt]{article}
\usepackage{coling2018}
\usepackage{times}
\usepackage{url}
\usepackage{latexsym}
\usepackage{pgfplots}
\usepackage{adjustbox}
\usepackage{amsmath}
\usepackage{multirow}
\usepackage{soul}

\pgfplotsset{compat=1.14}

\usepackage[disable]{todonotes}   

\newcommand{\note}[4][]{\todo[author=#2,color=#3,size=\scriptsize,fancyline,caption={},#1]{#4}} 

\newcommand{\manuel}[2][]{\note[#1]{Manuel}{orange!40}{#2}}

\newcommand{\katharina}[2][]{\note[#1]{Katharina}{yellow!40}{#2}}
\newcommand{\alfonso}[2][]{\note[#1]{Alfonso}{purple!40}{#2}}

\title{Analysis of Information Loss in Machine Translation Between Polysynthetic and Fusional Languages}
\title{Lost in Translation: Analysis of Information Loss During \\Machine Translation Between Polysynthetic and Fusional Languages}

\author{\parbox{\linewidth}{\centering
Manuel Mager{\rm\affmark[1]}, Elisabeth Mager{\rm\affmark[2]}, \\ 
Alfonso Medina-Urrea{\rm\affmark[3]}, Ivan Meza{\rm\affmark[1]},
Katharina Kann{\rm\affmark[4]}} \vspace{.12cm}
\\
\affaddr{\affmark[1]Instituto de Investigaciones en Matem\'aticas Aplicadas y en Sistemas,\\ Universidad Nacional Aut\'onoma de M\'exico, M\'exico}\\
\affaddr{\affmark[2]Facultad de Estudios Superiores Acatl\'an, Universidad Nacional Aut\'onoma de M\'exico, M\'exico}\\
\affaddr{\affmark[3]Centro de Estudios Ling\"u\'isticos y Literarios, El Colegio de M\'exico, M\'exico} \vspace{.1cm}\\
\affaddr{\affmark[4]Center for Data Science, New York University, USA}\\
\affaddr{\texttt{mmager@turing.iimas.unam.mx}}} 
\newcommand*{\affaddr}[1]{#1} 
\newcommand*{\affmark}[1][*]{\textsuperscript{#1}}

\date{}

\begin{document}
\maketitle

\vspace{40px}
\blfootnote{
    %
    %
    %
    %
    %
    %
     \hspace{-0.65cm}  
     This work is licensed under a Creative Commons 
     Attribution 4.0 International License.
     License details:
     \url{http://creativecommons.org/licenses/by/4.0/}.
}

\begin{abstract}
Machine translation from polysynthetic to fusional languages is a challenging task, which gets further complicated by the limited amount of parallel text available. 
Thus, translation performance is far from the state of the art for high-resource and more intensively studied language pairs. To shed light on the phenomena which hamper automatic translation to and from polysynthetic languages, we study translations from  three low-resource, polysynthetic languages (Nahuatl, Wixarika and Yorem Nokki) into Spanish and vice versa. Doing so, we find that in a morpheme-to-morpheme alignment an important amount of information contained in polysynthetic morphemes has no Spanish counterpart, and its translation is often omitted. We further conduct a qualitative analysis and, thus, identify morpheme types that are commonly hard to align or ignored in the translation process.
\end{abstract}

\section{Introduction}
\label{introduction:ref}

Until a few years ago, research on machine translation (MT) between polysynthetic and fusional languages did not get much attention from the natural language processing (NLP) community. 
Furthermore, with the rise of neural MT (NMT), the common assumption 
that machine learning approaches for MT were language independent routed the efforts into the direction of general model improvements. But this assumption does not hold completely  true, 
and, recently, efforts have been made to adapt models to individual languages, e.g., in order to improve poor results on morphologically-rich languages \cite{ataman2018compositional,al2014unsupervised,lee2016fully}. 
\newcite{koehn2005europarl} mentioned this problem while he analyzed the Europarl corpus, stating that ``translating from an information-rich into an information-poor language is easier than the other way around''. However, doing so, we unfortunately note a loss of information. This idea that some languages encode more information in one phrase than others given rise to many questions in linguistics 
and NLP, and motivated this paper. Polysynthetic languages are a special type of information-rich languages, and come with their own set of challenges for translation. Studying their particularities is an important prerequisite to enable successful translation to or from them in the future.

Many polysynthetic languages---many of which endangered---are spoken in regions where Spanish, English, or Portuguese are dominant. Thus, improving the translation quality of MT between fusional and polysynthetic languages might play an important role for communities which speak a polysynthetic language, e.g., by making documents in key fields such as legal, health and education accessible to them. Although many members of these communities can obtain access to this information using another dominant language which they also speak, this situation might have a negative effect on their native languages due to them not playing a functional role in day-to-day interaction about these important fields. As a result, these dominated languages might be perceived as less important. Well-performing MT might offer the mechanism to invert this situation, by making 
important documents accessible to the communities in their native languages, 
thus mitigating the need to consider one language more important, since both allow access to the same sets of documents. 
Rule-based MT (RBMT) has been a common approach to deal with low-resource MT. However, statistical MT (SMT) and NMT are essential for a broad coverage, due to the vast diversity of polysynthetic languages.  

In this paper, we introduce the following research questions
: (i) Which information is commonly not encoded in the target text when translating to a fusional language from a polysynthetic one? (ii) How can this information loss be explained from a linguistic point of view? (iii) Are some morphemes 
particularly hard to translate? 

In an attempt to start answering the before mentioned questions,
we present a quantitative study, using morpheme-based SMT alignments \cite{brown1993mathematics} between the following language pairs: Nahuatl-Spanish, Wixarika-Spanish, and Yorem Nokki-Spanish\footnote{The language we call Wixarika is also known as Huichol, and Yorem Nokki is also known as Mayo or Yaqui. Similarly, the Yuto-Nahua linguistic family also goes by the name of Uto-Aztec. We use these names out of respect to the communities that have chosen these names within the language. 
}. With the exception of Spanish, all these languages are from the Yuto-Nahua  linguistic family and have different levels of polysynthesis. We search for commonly not aligned morphemes 
and analyze the results. 
Trying to find answers to our research questions, we also present the qualitative aspects of this information loss.

\section{About Polysynthetic Languages
}
\label{alignment:ref}
Translating from a polysynthetic language to a fusional one faces difficulties; a significant number of morphemes can get lost because polysynthetic languages have structures that are different from those of fusional languages.
A main difference between the fusional and the polysynthetic languages lays at the syntax level of a sentence. Johanna Nichols refers to a binary system, ``directed relations between a head and a dependent'' \cite{nichols1986head}, between a head-marked and a dependent-marked relation. While the dependent-marked sentence is characterized by a relation of dependent pronoun-noun and relative construction, the head-marked construction prefers governed arguments, possessed noun, main-clause predicate and inner and outer adverbial constructions. For her ``the head is the word which governs, or is subcategorized for --or otherwise determines the possibility of occurrence of-- the other word. It determines the category of its phrase.''  In contrast of the most Indo-European languages,  ``the  Mayan, Athabaskan, Wakashan, Salishan, Iroquoian, Siouan, and Algonkian families are consistently head-marking'' \cite{nichols1986head}, among others.  So the polysynthetic languages prefer a head-marked morphology, where the verb has a preference position; ``the verb itself normally constitutes a complete sentence; full NP's are included only for emphasis, focus, disambiguation etc.'' \cite{nichols1986head}. 

In the same way, Baker distinguishes ``head-marking'' languages from ``dependent-marking'' ones \cite{baker1996polysynthesis}. The first type of languages, in the most cases, has a ``head-final structure (SOV)''  or a free word order, while the second type exhibits a head-initial structure (SVO or VSO) \cite{baker1996polysynthesis}. We must note that not all polysynthetic languages have the head-final structure, for example in Nahuatl we also find an SVO structure as we will see in \S\ref{sec:infloss}. 
Jeff MacSwan pointed out that in Southeast Puebla Nahuatl we can find different structures, depending the meaning of each sentence: the SVO-structure is the most natural, the VSO-structure is employed for focus and contrast only and the SOV-structure for light emphasis, but is ``also possible for focus or contrast'' \cite{macswan1998argument}. In contrast, in Wixarika the SOV-structure dominates, which makes it less flexible that Nahuatl. The same phenomenon can also be observed in fusional languages. While English prefers an SVO-structure, in German different orders are possible: we can find the SOV-structure only in subordinate sentences, while in main sentences we can have either an SVO-structure, or an OVS-structure for emphasis. In contrast, in Nahuatl the OVS-structure is unacceptable \cite{macswan1998argument}. 

Baker  states ``that in a polysynthetic language like Mohawk, all verbs necessarily agree with subjects, objects, and indirect objects, except for the special case when the direct object is incorporated into the verb'' \cite{baker1996polysynthesis}. So ``every argument of a head element must be related to a morpheme in the word containing that head (an agreement morpheme, or an incorporated root)'' \cite{baker1996polysynthesis}, often expressed by asserts. 
Jos\'{e} Luis Iturrioz and Paula G\'{o}mez L\'{o}pez observe a semantic relation between the predicate and the arguments. For this reason, the enunciative functions as a perspective or situation, individuation, or identification (attribution, reference), discursive cohesion, culminating in the integration of one clause in another \cite{leza2006gramatica}. This phenomenon cannot be observed in fusional languages; thus, morphemes with specific incorporation functions do not exist in fusional languages. Therefore translation of such kind of morphemes can be challenging in the machine translation process. Moreover, it can be difficult for these morphemes to be inferred when the target language is a polysynthetic one. 

Incorporation is a common phenomenon in polysynthetic languages. Wilhelm von Humboldt first described it. For him, incorporation has a syntactical, but not a morphological function \cite{leza2006gramatica}. In contrast, Marianne Mithun referred to noun incorporation (NI) \cite{mithun1986nature} in the context of verb morphology. For her ``New topics may be introduced in other ways, however IN's do, on occasion, serve to introduce new topics -- simply because they are parts of complex verbs denoting conceptually unitary activities'' \cite{mithun1986nature}. Also, the NI is not simply a combination of a noun with a verb stem ``to yield a more specific, derived verb stem'' \cite{mithun1986nature}. According to Baker's theory of incorporation, in polysynthetic languages there exists an interrelation of morphemes, in the way ``that one part of a derived stem is the syntactic complement of the other part. In both cases, syntactic argument relationships are being expressed morphologically'' \cite{baker1996polysynthesis}. Then, we have agreement morphemes, expressing the argument of the verb, e.g., pronominal affixes and incorporated roots \cite{baker1996polysynthesis}. ``The word-internal structure in these languages is very configurational indeed (...) the order of basic morphemes is also quite consistent (...) Furthermore, (...) this morpheme order provides a clue to the basic syntactic structure of these languages" \cite{baker1996polysynthesis}. 

One important property of polysynthetic languages is the high number of morphemes which often occur in the verb structure. According to Paula G\'{o}mez, in the Wixarika language there are three positions before the verb stem for approximately twelve prefixes, which correspond to different senses; for example, expression of localization, individuation, participation, aspects and modes of action, among others \cite{gomez1999huichol}. Like Nichols, his phenomenon of  three-place verbs we observe also in the  Bantu  and Mayan languages \cite{nichols1986head}. Hence, a considerable number of words, encoding a significant amount of information, may be produced when combining these items. \alfonso{Quit\'{e} lo de la polisemia y la polifuncionalidad. Con hablar de las muchas combinaciones posibles es suficiente.} \manuel{Me parece buena idea. No nos aportaba mucho al argumento y si nos complicaba. Gracias!}
In polysynthetic languages, there is a higher fragmentation of words and an interrelation of the morphemes, making their translation more difficult.  Nichols mentions that ``in a number of North American families (Uto-Aztecan, Yuman, Pomoan, Siouan, Algonkian, Cadoan), instrumental, locative, and directional affixes on verbs are grammaticalized'' \cite{nichols1986head}. Besides, the morphology of the Wixarika verbs presents complications due to the great number of positions of affixes, which can reach to 20 morphemes or more, forming a morphological chain \cite{gomez1999huichol}. Also, in Wixarika,  the spatial or local relationships are expressed by adverbs, postpositions, nominal suffixes, and verbal prefixes \cite{gomez1999huichol}. For this reason, in many cases, we do not have a correlation of structures in a pair consisting of a polysynthetic and a fusional language, which can make translating difficult. 
In languages like Spanish, this information is commonly inferred or not encoded in the translation between languages of these two typologies. For instance, consider the Wixarika morpheme ``u'' that indicates that an action happens in the visual sphere of the speaker. This information is usually not directly translated to Spanish. However, as a result, a problem arises when trying to translate such a phrase back to Wixarika, because the information if the action is held in the visual field of the speaker or not is not available.  

\section{Previous Work}
Nowadays, the area of NLP is largely dominated by data-driven---often neural---approaches which, as far as the machine learning model is concerned, strive to be language-independent. 
However, the performance of such systems does still vary in dependence of the typology of each language. In order to shed light on this phenomenon and its causes, \newcite{cotterell-et-al-2018-lm} studied how  the difficulty of language modeling (LM) depends on the language. They found inflectional morphology to be a key factor: even character-based long short-term memory (LSTM) models performed worse for morphologically rich languages.  In this paper, we will study the causes of possible performance loss for polysynthetic languages in MT. 

MT has made a big step forward with the development of SMT and, later on, NMT. 
Those approaches make it possible to construct reasonably well performing MT systems using parallel corpora that can be gathered from a wide range of sources, and do not need 
handwritten rules generated by experts. 
This is crucial, due to the large number of polysynthetic languages. However, most systems fail to achieve good performances for polysynthetic languages, particularly in a low-resource context; with often better results for SMT than for NMT \cite{mager2018dh}. In recent years, character-based NMT systems were claimed to handle sub-word phenomena \cite{sennrich2016neural}, and others target specifically morphological rich languages \cite{passban2018improving,moprhnmt2018coling}, but with the condition of feeding the neural network with vast amounts of data. Character-based NMT systems can learn some morphological aspects \cite{belinkov2017neural} even for morphologically rich languages like Arabic. \newcite{P17-1184} analyzed this proposal, concluding that, although character-based models improve translation, best results are achieved with accurate morphological segmentation. 

The problem exposed in this paper has mainly been studied in the context of SMT approaches. This line of research has pursued both the goal of improving translation from morphologically-rich languages into morphologically-poorer ones like English \cite{habash2006arabic}, and the other way around \cite{avramidis2008enriching,oflazer2008statistical}. One important development was the inclusion of linguistic markups into factored translation models \cite{koehn2007factored,oflazer2008statistical,fraser2009experiments}. \newcite{virpioja2007morphology} proposed a combined usage of Morfessor \cite{creutz2005unsupervised}, an unsupervised segmentation model, and phrase-based SMT systems, in order to make use of segmented input. The translation improvement through initial morphological segmentation 
was also found
for translation of the polysynthetic Wixarika into Spanish \cite{mager2016traductor}. In each case the main goal of previous work was to increase the BLEU score. However, in this paper we aim to improve our understanding  of the information which is lost in the translation process; and particularly for polysynthetic languages. For this, we make use of morphological segmentation.

\section{Morpheme Alignment Between a Polysynthetic and a Fusional Language}
Morphemes are the smallest meaning-bearing units of words. Here, we want to know why some of them are not aligned correctly 
by common SMT systems. We use the surface form of morphemes obtained from manually segmented words 
in each language and apply the IBM models 2, 3 and 4 \cite{brown1993mathematics} to get alignment cepts for each morpheme. Each cept contains a (possibly empty) set of positions with which a token is aligned. Also a special cept is defined, that is aligned to morphemes that could not be aligned to any other cept, and is referred to with number 0. \katharina{This sentence is a bit strange.} A set of alignments is denoted by 
 $A(e, f)$, where $f$ is a phrase of size $m$ in a source language and $e$ is a phrase of size $l$ in a target language.

For our experiments, we use the resulting alignment function obtained from the training process which consists of maximizing the translation likelihood of the training set.
The probability of a translation $f$ in the target language, given a source sentence $e$, is then calculated as:
\begin{align}
Pr(f|e) &= \sum_a Pr(f,a|e)\label{trans}
\end{align}

  The alignments $a \in A$ are trained jointly with the whole translation model, as defined by \newcite{brown1993mathematics}. 
  The underlined conditional probability $a$ in Equation \ref{eq:cond} is what we use for our work.
\begin{align}
Pr(f,a|e) &= P(m|e)\prod_{j=1}^{m}\underbrace{Pr(a_j|a^{j-1}_1,f^{j-1}_1,m,e)}_a Pr(f_j|a^j_1,f^{j-1}_1,m,e) \label{eq:cond}
\end{align}

After computing the alignments for each sentence pair in our dataset, we count all morphemes which are not aligned, in order to find information that is not expressed in the target language. As SMT is a probabilistic method the resulting alienations are not exact and should be taken as an approximation. Adding this fact, also the amount of data used to train the system influences the resulting inference. 

\section{Experiments}
\label{sec:experiments}
In order to get the alignments of morphemes in Nahuatl, Wixarika and Yorem Nokki with their Spanish counterparts, 
we train a word-based SMT system using GIZA++ \cite{och03:asc} on  parallel datasets in the respective languages, which we will describe below. We parsed the resulting alignment files to extract non-aligned morphemes in each translation direction. As we are working with three language pairs and are interested in both translation directions, we trained six models (Spanish-Wixarika, Wixarika-Spanish, Spanish-Nahuatl, Nahuatl-Spanish, Spanish-Yorem Nokki and Yorem Nokki-Spanish).

\subsection{Languages}

\paragraph{Nahuatl}
 is a language of the Yuto-Nahua language family and the most spoken native language in Mexico. The variants of this language are diverse, and in some cases can be considered linguistic subgroups. Its almost $1,700$ thousand native speakers mainly live in Puebla, Guerrero, Hidalgo, Veracruz, and San Luis Potosi, but also in Oaxaca, Durango, Modelos, Mexico City, Tlaxcala, Michoacan, Nayarit and the State of Mexico. In this work we use the Northern Puebla variant.

Like all languages of the Yuto-Nahua family, Nahuatl is agglutinative, and a word can consist of a combination of  many different morphemes. In contrast to other languages, Nahuatl 
has a preference for SVO, but SOV and VSO are also used. An example phrase is:

\begin{center}

  \begin{tabular}{ l  l  l }
ka-te &  ome & no-kni-wan \\
is-pl. & two & pos.1sg:s.-brother-pl. 
  \end{tabular}
  
\emph{I have two brothers} 
\end{center}

This Nahuatl variant has five vowels ($\{i,u,e,o,a\}$) and does not distinguish if they are short or large.  The alphabet of Nahuatl consists of 17 symbols:  $\Sigma_{Nahuatl}$ = \textit{\{a,e,h,i,k,m,n,o,p,r,s,t,u,w,x,y,'\}}. 

\paragraph{Wixarika} 
is a language spoken by approximately fifty thousand people in the Mexican states of Jalisco, Nayarit, Durango and Zacatecas. It belongs to the Coracholan group of languages within the Yuto-Nahua family. Its alphabet 
consists of 17 symbols ($\Sigma_{Wixarika}$=\textit{\{`,a,e,h,i,+,k,m,n,p,r,t,s,u,w,x,y\}}), out of which  five are vowels \{a,e,i,u,+\}\footnote{While linguists often use a dashed i (\st{i}) to denote the last vowel, in practice almost all native speakers use a plus symbol (+). In this work, we chose to use the latter.} with long and short variants.
An example for a phrase in the language is:
\begin{center}

  \begin{tabular}{ l  l }
yu-huta-me & ne-p+-we-'iwa \\
an-two-ns & 1sg:s-asi-2pl:o-brother 
  \end{tabular}
  
\emph{I have two brothers} 
\end{center}

~\\
This language has a strong SOV syntax, with heavy agglutination on the verb. Wixarika is considered morphologically more complex than other languages from the same family \cite{leza2006gramatica}.   

\paragraph{Yorem Nokki} 
 is part of the Taracachita subgroup of the Yuto-Nuahuan language family.
Its Southern dialect (commonly known as Mayo) is spoken by close to fifty thousand people in the Mexican states of Sinaloa and Durango, 
while its Northern dialect (also known as Yaqui) has about forty thousand speakers in the Mexican state of Sonora.
In this work, we consider the Mayo dialect.
As in the other studied languages, the nominal morphology of Yorem Nokki is rather simple, but, again, 
the verb is highly complex. 
Yorem Nokki uses mostly SOV.

The symbols used in our dataset are $\Sigma_{YoremNokki}$= \textit{\{\'{a},k,s,g,\'{o},j,y,w,$\beta$,p,m,e,n,d,r,\'{e},t,u,c,o,h,f,b,',i,l,a\}}, with 8 being vowels: \{a,e,i,o,u,\'{a},\'{e},\'{o}\}. An example phrase is:

\begin{center}

  \begin{tabular}{ l l l }
  woori&saila-m-ne&hipu-re \\
  two&brother-pl.-me&have-r
  \end{tabular}
  
\emph{I have two brothers} 
\end{center}

\begin{center}
\end{center}

\subsection{Dataset}\label{sec:dataset} 
To create our datasets, we use phrases in the indigenous languages paired with their Spanish equivalents. Namely, our translations are taken from books of the collection \textit{Archive of Indigenous Languages} for Nahuatl \cite{lastra1980nahuatl}, Wixarika \cite{gomez1999huichol}, and Yorem Nokki \cite{freeze1989mayo}. A total of 594 phrases is available for each language. To obtain these phrases, a questionnaire of 594 utterances made by Ray Freeze was used \cite[p. 15]{freeze1989mayo}. In essence, each author elicited equivalent expressions from speakers of the target language. Also, the uniformity of the questionnaire may have been modified because of cultural or environmental circumstances. In cases where the expression could not be elicited, a sentence was offered as similar as possible, grammatically and semantically, to the original utterance. In this manner, the sets of expressions are equivalent for all languages, so we can directly compare results\alfonso{Dec\'{\i}a "the sets of translations are equivalent for all languages, so we can directly compare all results"; lo matic\'{e} un poco}\manuel{Quedó genial}. The words of the polysynthetic languages have already been segmented by linguists in the cited books. In order to achieve a morpheme-to-morpheme translation (instead of a word-to-morpheme one) we segment the Spanish phrases with Morfessor \cite{virpioja2013morfessor} and manually correct segmentation errors.


\begin{table}[!h]
    \begin{center}
    \small
        \begin{tabular}{|l||r|r|r||l|r|r|r|}
                \hline
\multicolumn{4}{|c|}{\bf Wixarika - Spanish} &\multicolumn{4}{|c|}{\bf Spanish - Wixarika} \\
        \hline
            \hline \bf Token & \bf Alig & \bf Non & \bf Diff & \bf Token & \bf Alig & \bf Non & \bf Diff\\\hline
p+- & 4 & 294 & -290&   el & 13 & 119 & -106\\
ne- & 34 & 208 & -174&  de & 13 & 81 & -68\\
ti- & 3 & 162 & -159&   -s & 33 & 86 & -53\\
p-& 24 & 153 & -129&   en & 7 & 58 & -51\\
u-& 2 & 102 & -100&    ?` & 8 & 47 & -39\\
a-& 7 & 106 & -99&     la & 22 & 61 & -39\\
e-& 3 & 74 & -71&      que & 27 & 52 & -25\\
r-& 9 & 73 & -64&      a & 46 & 65 & -19\\
eu- & 2 & 64 & -62&     ! & 5 & 19 & -14\\
t+a & 3 & 54 & -51&           con & 5 & 19 & -14\\
k+ & 1 & 47 & -46&            lo & 6 & 17 & -11\\
ta- & 9 & 52 & -43&     ? & 24 & 33 & -9\\
m+- & 6 & 43 & -37&     \'{o} & 5 & 13 & -8\\
'u- & 3 & 37 & -34&     \'{e}s & 0 & 6 & -6\\
ye- & 12 & 44 & -32&    casa & 11 & 17 & -6\\
\hline
\hline
\multicolumn{4}{|c|}{\bf Yorem Nokki - Spanish} &\multicolumn{4}{|c|}{\bf Spanish - Yorem Nokki} \\
\hline
\hline \bf Token & \bf Alig & \bf Non & \bf Diff & \bf Token & \bf Alig & \bf Non & \bf Diff\\\hline
-k & 1 & 185 & -184&         est\'{a} & 6 & 37 & -31\\
-ka & 25 & 143 & -118&       ? & 11 & 32 & -21\\
ta & 8 & 88 & -80&          la & 32 & 48 & -16\\
ne & 40 & 114 & -74&        con & 7 & 17 & -10\\
si & 1 & 58 & -57&          para & 4 & 13 & -9\\
e' & 7 & 62 & -55&          que & 33 & 41 & -8\\
wa & 1 & 54 & -53&          va & 5 & 13 & -8\\
wa- & 2 & 55 & -53&   	    -ndo & 1 & 9 & -8\\
a & 25 & 72 & -47&          al & 8 & 16 & -8\\
ka- & 6 & 52 & -46&   se & 25 & 33 & -8\\
wi & 7 & 49 & -42&          un & 1 & 8 & -7\\
a' & 5 & 46 & -41&          yo & 7 & 12 & -5\\
po & 36 & 73 & -37&         las & 3 & 8 & -5\\    
$\beta$a &3&37&-37&			est\'an&3&8&-5\\
ta-&5&35&-30&			ustedes&3&8&-5\\
\hline
\hline
\multicolumn{4}{|c|}{\bf Nahuatl - Spanish} &\multicolumn{4}{|c|}{\bf Spanish - Nahuatl} \\
\hline
            \hline \bf Token & \bf Alig & \bf Non & \bf Diff & \bf Token & \bf Alig & \bf Non & \bf Diff\\\hline
o-&25 & 214 & -189     -a & 10 & 95 & -85\\
in & 77 & 222 & -145&         -s & 30 & 82 & -52\\
-tl & 2 & 108 & -106&   ? & 6 & 45 & -39\\ 
ni- & 9 & 111 & -102&   es & 18 & 45 & -27 \\
i & 23 & 73 & -50&            de & 32 & 55 & -23\\ 
k- & 6 & 54 & -48&      ! & 3 & 21 & -18\\ 
ki- & 11 & 54 & -43&    que & 29 & 45 & -16\\ 
ti- & 13 & 56 & -43&    ?` & 17 & 32 & -15\\  
mo- & 16 & 54 & -38&    est\'{a} & 15 & 22 & -7\\
n- & 2 & 40 & -38&      -o & 1 & 7 & -6\\ 
k & 10 & 37 & -27&            -do & 3 & 8 & -5\\ 
'ke & 5 & 29 & -24&           -n & 1 & 6 & -5\\ 
te & 3 & 25 & -22&            con & 10 & 14 & -4\\   
ka- & 10 & 29 & -19&    est\'{a}n & 2 & 6 & -4\\     
-to & 4 & 22 & -22 &    -ra & 0 & 4 & -4 \\
                             
\hline
\end{tabular}
    \end{center}
    \caption{\label{tab:wix} Alignment results between language pairs. The \textit{Token} column stands for a word or morpheme (morphemes contains the - symbol), \textit{Alig} is the number of times that the token was aligned, \textit{Non} is the number of times the token was not aligned, and \textit{Diff} is the difference between the numbers of aligned and non-aligned tokens.}
\end{table}

\subsection{Results and Discussion}
Table \ref{tab:wix} shows the top fifteen non-aligned morphemes resulting from translating Nahuatl, Wixarika, and Yorem Nokki into Spanish. Naturally, it would be interesting to discuss the syntax of all morphemes in detail. In this way it could be established, for example, what kinds of markers are more characteristic of which kind of language. However, 
this is out-of-scope for this work, such that
we will limit ourselves to the following remarks.\alfonso{ojal\'{a} esto sirva de vacuna al deseo del dictaminador de convertir este trabajo en otra cosa} 

For Wixarika we can see that the eight morphemes which are most challenging 
for translation are sub-word units.  
The important Wixarika independent asserters ``p+'' and ``p'' are the most frequent morphemes in this language. However, as they have no direct equivalent in Spanish, their translation is mostly ignored. The same is true for the object agreement morphemes ``a'' and ``ne''. This is particularly problematic for the translation in the other direction, i.e., from Spanish into Wixarika, as a translator has no information about how the target language should realize such constructions. Human translators can, in some cases, infer the missing  information. However, without context it is generally complicated to get the right translation. 
Also, other morphemes like ``u'', ``e'', and ``r'' encode precise information about forms and movement that are not usually expressed in Spanish. Some of the other morphemes for which the alignment fails, like ``ne'' and ``ti'', could have been translated as a first person possessive, or as a question mark, respectively. The reason for those errors might be our low-resource setting. 

The Yorem Nokki suffixes ``k'' and ``ka'' are realization morphemes. As this construction is not commonly expressed in a fusional or isolating language, it will frequently not be aligned with any Spanish token.  Another difficulty is the translation of a concatenative word construction into a fusional one. Yorem Nokki does not use as many agreement morphemes as Wixarika or Nahuatl, but the ``si'' noun agreement morpheme still is one of the hardest to align. The morpheme ``ta'' represents the accusative verbal case that can be expressed in Spanish, but still appears as one of the most difficult morphemes to align. This can be a consequence of the low-resource setting we have. 

The last language pair (Nahuatl-Spanish) has less non-aligned morphemes than the previews ones. Nahuatl is also the only language for which the most frequently unaligned token is a word: the token ``in'' is an article. However, its translation is not trivial. As most Yuto-Nahua languages, Nahuatl does not mark grammatical gender \cite{mager:2017}. The lack of gender information can hurt SMT performance. In practice, for such cases post-processing can be used to correct some of the system errors \cite{stymne2011pre}. However, as in the previous cases, the object (``k'' and ``ki'') a noun agreement morphemes (``ni'' and ``ti'') are the most frequently unaligned morphemes. 

Table \ref{tab:stats} shows the amount of non-aligned tokens, words, and morphemes for each language pair and each translation direction. Our first observation is that the rate of non-aligned tokens for the direction Spanish-polysynthetic language is far lower than the other way around. The highest rates of non-aligned tokens are found for Wixarika and Yorem Nokki with $0.617$ and $0.616$, respectively. For Nahuatl and Spanish, this rate is with $0.448$ notably lower. On the other hand, the translation direction from Spanish to our polysynthetic languages seems to work much better and shows less variability. The lowest rate is obtained for Yorem Nokki with $0.264$, followed by Nahuatl with $0.277$, and Wixarika with $0.35$. 
For both directions, translation with Wixarika got the highest non-alignment rates. 
This suggests that the phenomenon might be related to the number of morphemes per word: 
\newcite{kann2018} showed that Wixarika has the highest morphemes-per-word rate among the languages considered here.  

\begin{table}[ht]
    \begin{center}
    \small
        \begin{tabular}{|l||r|r|r|r|r|}
                \hline
&\bf Tokens & \bf N.a. Tokens & \bf N.a. words & \bf N.a. morph. & \bf N.a./tokens \\
\hline
\bf Wixarika-Spanish & 4702& 2905& 790& 2115& 0.617 \\
\bf Spanish-Wixarika & 3594& 1259& 1259& 0& 0.350 \\
\bf Nahuatl-Spanish & 4391& 1969& 1111& 858 & 0.448 \\
\bf Spanish-Nahuatl & 3380 & 939 & 939&0 &0.277 \\
\bf Yorem Nokki-Spanish & 4805& 2960 & 2238 & 722 & 0.616  \\
\bf Spanish-Yorem Nokki & 3163 & 836 & 836 & 0 & 0.264 \\
        \hline
\end{tabular}
    \end{center}
    \caption{\label{tab:stats} Alignments of tokens, words, morphemes and their success rates for all language pairs. \textit{N.a. Tokens} counts all non-aligned tokes, \textit{N.a. words} counts only non-aligned words (one morpheme per token), \textit{N.a. morph.} are the non-aligned morphemes. \textit{N.a./tokens} is the rate of non-aligned tokens in relation to the total number of tokens.}
\end{table}

To sum up, some fine-grained information from verbs in our polysynthetic languages are not usually translated to Spanish, since this information is not commonly expressed in this target languages. It is particularly true for structure morphemes and agreement morphemes, as well as enunciation functions (situations, individuation, attribution, reference, and discursive cohesion).  

The severity of the alignment problem seems to correlate with the BLEU metric for translation of these language pairs \cite{mager2018dh}: translation from polysynthetic languages to fusional ones has been reported to work notably better than the opposite direction. We expect the alignment issue to be fundamental for explaining this dynamic.

\section{Information Loss in Translation Between Polysynthetic and Fusional Languages}
\label{sec:infloss}

As discussed in \S\ref{sec:experiments}, an important amount of morphemes from a polysynthetic language usually will often not be aligned to morphemes of an fusional language. However, how can we explain such a loss of information? In this section, 
we will conduct a qualitative analysis to obtain a better understanding of this phenomenon.  

Namely, we will analyze the phrase ``She always asks us for tortillas.'' in our three polysynthetic languages.  The first example, taken from \newcite{gomez1999huichol}, will be in Wixarika:
\begin{center}

\begin{tabular}{l l l l}
m+k+&pa:pa&ya&p+-ta-ti-u-ti-wawi-ri-wa\\
Ella&tortilla&enf&asi-1pl:o-its-vis-pl:a-pedir-apl-hab \\
\end{tabular}
\begin{tabular}{r l}
\textit{Free Spanish translation}:&Ella siempre nos pide tortillas\\
\textit{Free English translation}:&She always asks us for tortillas\\Abbreviations: & \\
enf      &     empathic \\
asi       &      independent asserter \\
pl:o     &      plural of the indirect object \\
its        &     intensifier \\
vis       &     visible: in the ambit of the  speaker\\
pl:a     &     plurality of the action\\
appl    &    applicative\\
hab    &    habitual
\end{tabular}
\end{center}
 
Wixarika employs a head-final structure (SOV) as can be seen in the example. 
Therefore, we have in the third place the  emphatic factor ``ya'' which realizes the agreement of the initial subject and the direct object;  also we need an asserter for the indirect object ``ta'', what in this case is the morpheme ``p+'', which we cannot translate. In fact, ``p+'' was the most unaligned morpheme in table \ref{tab:wix}. The verb exists of different prefixes collocated before the verb stem ``wawi'': the morpheme ``ti'' is an intensifier of the visibility of the ambit of the speaker, expressed with the morpheme ``u''; the prefix ``ti'' on the first place of the verb refers to the plurality of the action and the plural of the direct object. Therefore, we can speak of an incorporation of the object into the verb.  All these prefixes cannot be directly translated.

Next, we analyze our example phrase in the Nahuatl Acaxochitlan dialect, spoken in Hidalgo \cite{lastra1980nahuatl}, and its translation into Spanish: 

\begin{center}
\begin{tabular}{l l l l}
ye'wa&tech-tla-tlanilia&semian&in-tlaxkal-i\\
ella&1pl.obj-indef-pide&siempre&art-tortilla-abs\\
\end{tabular}

\begin{tabular}{r  l}
\textit{Free Spanish translation}:   &           Ella siempre nos pide tortillas\\
\textit{Free English translation}:&She always asks us for tortillas\\
Abbreviations: & \\
abs      &        absolutive \\
art        &      article \\
indef    &        indefinite\\
obj       &        object \\
pl         &         plural
\end{tabular}
\end{center}

In the Nahuatl language we also see a loss of information, but less than in Wixarika. Here, we notice a different syntactic structure: the direct object is located at the end of the phrase and the indirect object is located in the second place. Thus, we have an SOVO structure. What is not translated is the prefix ``ya'' of the verb ``yanilia'' (ask for) because the undefined situation of an action is unknown in the fusional languages, as well as the absolutive suffix ``i'' of the object ``tlaxkal'' (tortillas)

Finally, we consider the same example for Yorem Nokki, taken from the Mayo dialect of Yorem Nokki (Southern branch), which is spoken in the Mexican state of Sinaloa \cite{freeze1989mayo}:
\begin{center}
\begin{tabular}{l l l l l} 
hi$\beta$a&a:po&tahkari-m&ito-wi&a'a:wa \\
siempre&ella &ortilla-pl&nosotros-a&hab-pide\\
\end{tabular}
\begin{tabular}{r  l}
\textit{Free Spanish translation}:     &     Ella siempre nos pide tortillas \\
\textit{Free English translation}:&She always asks us for tortillas\\
Abbreviations: & \\
hab    &     habitual \\
pl         &         plural
\end{tabular}
\end{center}

In the Yorem Nokki language, we have a head final structure (SOV) like in Wixarika. However this phrase has only one morpheme that cannot be directly translated to Spanish, e.g., the prefix ``a'' that expresses an habitual action.

\section{Conclusion and Future Work}

We presented a quantitative and a qualitative study of the information loss that occurs during MT from three polysynthetic languages of the Yuto-Nahua 
family into Spanish, a fusional language, and vice versa. Based on GIZA++ alignments between Spanish morphemes and the corresponding morphemes in our polysynthetic languages, we got insight into which morphemes are commonly not translated. We found that, in contrast to the morphemes in the polysynthetic languages, most Spanish tokens get aligned by the aligner. 

We further noticed that often fine-grained information which is encoded into polysynthetic verbs is not translated, since this information is not commonly expressed in our fusional language. The same holds true for polysynthetic structure morphemes and agreement morphemes. Other morphemes which are hard to translate are the enunciation functions, like perspectives or situations, individuation, attribution, reference, and discursive cohesion. In Wixarika, the hardest morphemes to translate are the assignors ``p+'', ``p'' and ``m+'', the object agreement morphemes ``a'' and ``ne'', and the action perspective morphemes ``u'' and ``e''; for Yorem Nokki the realization morphemes ``ka'' and ``k'' and the ``si'' noun agreement morpheme; for Nahuatl the object agreement morphemes like ``k'' and ``ki'' and the noun agreement morphemes ``ni'' and ``ti''. By revision of non-aligned morphemes we could also see that our three analyzed polysynthetic languages have entirely different structures, but in all cases, the agreement morphemes represented were hard to align with Spanish morphemes.

In future work, 
we aim to increase the amount of data to train our MT models. For instance, with the usage of automatic morphological segmentation systems like the one presented by \newcite{kann2018}, we could use larger amounts of parallel data for training and, thus, reduce alignment errors in our experiments. Such an error reduction for alignments could help us to identify in a cleaner way the underlying phenomena that hurt MT for the languages considered here.

\section*{Acknowledgments}
We would like to thank all the anonymous reviewers for their valuable comments and feedback.

\bibliographystyle{acl}
\bibliography{acl2016}

\begin{thebibliography}{}

\bibitem[\protect\citename{Al-Mannai \bgroup et al.\egroup
  }2014]{al2014unsupervised}
Kamla Al-Mannai, Hassan Sajjad, Alaa Khader, Fahad Al~Obaidli, Preslav Nakov,
  and Stephan Vogel.
\newblock 2014.
\newblock Unsupervised word segmentation improves dialectal arabic to english
  machine translation.
\newblock In {\em Proceedings of the ANLP}, pages 207--216.

\bibitem[\protect\citename{Ataman and Federico}2018]{ataman2018compositional}
Duygu Ataman and Marcello Federico.
\newblock 2018.
\newblock Compositional representation of morphologically-rich input for neural
  machine translation.
\newblock {\em arXiv preprint arXiv:1805.02036}.

\bibitem[\protect\citename{Avramidis and Koehn}2008]{avramidis2008enriching}
Eleftherios Avramidis and Philipp Koehn.
\newblock 2008.
\newblock Enriching morphologically poor languages for statistical machine
  translation.
\newblock {\em Proceedings of 2008 Anual Meeting of ACL}, pages 763--770.

\bibitem[\protect\citename{Baker}1996]{baker1996polysynthesis}
Mark~C Baker.
\newblock 1996.
\newblock {\em The polysynthesis parameter}.
\newblock Oxford University Press.

\bibitem[\protect\citename{Belinkov \bgroup et al.\egroup
  }2017]{belinkov2017neural}
Yonatan Belinkov, Nadir Durrani, Fahim Dalvi, Hassan Sajjad, and James Glass.
\newblock 2017.
\newblock What do neural machine translation models learn about morphology?
\newblock In {\em Proceedings of the 2017 Annual Meeting of the ACL}, volume~1,
  pages 861--872.

\bibitem[\protect\citename{Brown \bgroup et al.\egroup
  }1993]{brown1993mathematics}
Peter~F Brown, Vincent J~Della Pietra, Stephen A~Della Pietra, and Robert~L
  Mercer.
\newblock 1993.
\newblock The mathematics of statistical machine translation: Parameter
  estimation.
\newblock {\em Computational linguistics}, 19(2):263--311.

\bibitem[\protect\citename{Cotterell \bgroup et al.\egroup
  }2018]{cotterell-et-al-2018-lm}
Ryan Cotterell, Sebastian~J. Mielke, Jason Eisner, and Brian Roark.
\newblock 2018.
\newblock Are all languages equally hard to language-model?
\newblock In {\em Proceedings of the 2018 Conference of the NAACL-HLT}, New
  Orleans, June.

\bibitem[\protect\citename{Creutz and Lagus}2005]{creutz2005unsupervised}
Mathias Creutz and Krista Lagus.
\newblock 2005.
\newblock {\em Unsupervised morpheme segmentation and morphology induction from
  text corpora using Morfessor 1.0}.
\newblock Helsinki University of Technology Helsinki.

\bibitem[\protect\citename{Fraser}2009]{fraser2009experiments}
Alexander Fraser.
\newblock 2009.
\newblock Experiments in morphosyntactic processing for translating to and from
  german.
\newblock In {\em Proceedings of the Fourth WMT}, pages 115--119. Association
  for Computational Linguistics.

\bibitem[\protect\citename{Freeze}1989]{freeze1989mayo}
Ray~A Freeze.
\newblock 1989.
\newblock Mayo de {L}os {C}apomos, {S}inaloa.
\newblock In {\em Archivos de lenguas ind{\'\i}genas de M{\'e}xico}, volume~14.
  El Colegio de M{\'e}xico.

\bibitem[\protect\citename{G{\'o}mez}1999]{gomez1999huichol}
Paula G{\'o}mez.
\newblock 1999.
\newblock Huichol de {S}an {A}ndr{\'e}s {C}ohamiata, {J}alisco.
\newblock In {\em Archivos de lenguas ind{\'\i}genas de M{\'e}xico}, volume~22.
  El Colegio de M{\'e}xico.

\bibitem[\protect\citename{Habash and Sadat}2006]{habash2006arabic}
Nizar Habash and Fatiha Sadat.
\newblock 2006.
\newblock Arabic preprocessing schemes for statistical machine translation.
\newblock In {\em Proceedings of the 2006 Conference of NAACL-HLT}, pages
  49--52. Association for Computational Linguistics.

\bibitem[\protect\citename{Iturrioz~Leza and
  L{\'o}pez~G{\'o}mez}2006]{leza2006gramatica}
Jos{\'e}~Luis Iturrioz~Leza and Paula L{\'o}pez~G{\'o}mez.
\newblock 2006.
\newblock {\em Gram{\'a}tica wixarika}, volume~1.
\newblock Lincom Europa.

\bibitem[\protect\citename{Kann \bgroup et al.\egroup }2018]{kann2018}
Katharina Kann, Manuel Mager, Ivan Meza, and Hinrich Sh\"{u}tze.
\newblock 2018.
\newblock Fortification of neural morphological segmentation models for
  polysynthetic minimal-resource languages.
\newblock In {\em Proceedings of the 2018 Conference of the NAACL-HLT 2018}.
  North American chapter of the Association for Computational Linguistics.

\bibitem[\protect\citename{Koehn and Hoang}2007]{koehn2007factored}
Philipp Koehn and Hieu Hoang.
\newblock 2007.
\newblock Factored translation models.
\newblock In {\em Proceedings of the 2007 joint conference EMNLP-CoNLL}.

\bibitem[\protect\citename{Koehn}2005]{koehn2005europarl}
Philipp Koehn.
\newblock 2005.
\newblock Europarl: A parallel corpus for statistical machine translation.
\newblock In {\em MT summit}, volume~5, pages 79--86.

\bibitem[\protect\citename{Lastra~de Su{\'a}rez}1980]{lastra1980nahuatl}
Yolanda Lastra~de Su{\'a}rez.
\newblock 1980.
\newblock N{\'a}huatl de {A}caxochitl{\'a}n, {H}idalgo.
\newblock In {\em Archivos de lenguas ind{\'\i}genas de M{\'e}xico}, volume~10.
  El Colegio de M{\'e}xico.

\bibitem[\protect\citename{Lee \bgroup et al.\egroup }2016]{lee2016fully}
Jason Lee, Kyunghyun Cho, and Thomas Hofmann.
\newblock 2016.
\newblock Fully character-level neural machine translation without explicit
  segmentation.
\newblock {\em arXiv preprint arXiv:1610.03017}.

\bibitem[\protect\citename{MacSwan}1998]{macswan1998argument}
Jeff MacSwan.
\newblock 1998.
\newblock The argument status of nps in southeast puebla nahuatl: comments on
  the polysynthesis parameter.
\newblock {\em Southwest Journal of Linguistics}, 17(2):101--114.

\bibitem[\protect\citename{Mager and Meza}2018]{mager2018dh}
Manuel Mager and Ivan Meza.
\newblock 2018.
\newblock Hacia la traducci\'{o}n autom\'{a}tica de las lenguas ind\'{i}genas
  de m\'{e}xico.
\newblock In {\em Proceedings of the DH 2018}. The Association of Digital
  Humanities Organizations.

\bibitem[\protect\citename{Mager~Hois \bgroup et al.\egroup
  }2016]{mager2016traductor}
Jes{\'u}s~Manuel Mager~Hois, Carlos Barr{\'o}n~Romero, and Ivan~Vladimir
  Meza~Ruiz.
\newblock 2016.
\newblock Traductor estad{\'\i}stico wixarika-espa{\~n}ol usando
  descomposici{\'o}n morfol{\'o}gica.
\newblock In {\em Proceedings of COMTEL}. Universidad Inca Garcilaso de la
  Vega.

\bibitem[\protect\citename{Mager~Hois}2017]{mager:2017}
Jesus~Manuel Mager~Hois.
\newblock 2017.
\newblock Traductor h\'{i}brido wix\'{a}rika - espa{\~n}ol con escasos recursos
  biling\"{u}es.
\newblock Master's thesis, Universidad Aut\'{o}noma Metropolitana.

\bibitem[\protect\citename{Mithun}1986]{mithun1986nature}
Marianne Mithun.
\newblock 1986.
\newblock On the nature of noun incorporation.
\newblock {\em Language}, 62(1):32--37.

\bibitem[\protect\citename{Nichols}1986]{nichols1986head}
J~Ohanna Nichols.
\newblock 1986.
\newblock Head-marking and dependent-marking grammar.
\newblock {\em Language}, 62(1):56--119.

\bibitem[\protect\citename{Och and Ney}2003]{och03:asc}
Franz~Josef Och and Hermann Ney.
\newblock 2003.
\newblock A systematic comparison of various statistical alignment models.
\newblock {\em Computational Linguistics}, 29(1):19--51.

\bibitem[\protect\citename{Oflazer}2008]{oflazer2008statistical}
Kemal Oflazer.
\newblock 2008.
\newblock Statistical machine translation into a morphologically complex
  language.
\newblock In {\em Proceeding of CICLing}, pages 376--387. Springer.

\bibitem[\protect\citename{Passban \bgroup et al.\egroup
  }2018]{passban2018improving}
Peyman Passban, Qun Liu, and Andy Way.
\newblock 2018.
\newblock Improving character-based decoding using target-side morphological
  information for neural machine translation.
\newblock In {\em Proceedings of the 2018 Conference of the NAACL-HLT},
  volume~1, pages 58--68.

\bibitem[\protect\citename{Peyman \bgroup et al.\egroup
  }2018]{moprhnmt2018coling}
Passban Peyman, Way Andy, and Liu Qun.
\newblock 2018.
\newblock Tailoring neural architectures for translating from morphologically
  rich languages.
\newblock In {\em Proceedings of the 2018 COLING)}. nternational Committee on
  Computational Linguistics.

\bibitem[\protect\citename{Sennrich \bgroup et al.\egroup
  }2016]{sennrich2016neural}
Rico Sennrich, Barry Haddow, and Alexandra Birch.
\newblock 2016.
\newblock Neural machine translation of rare words with subword units.
\newblock In {\em Proceedings of the 2016 Annual Meeting of the ACL}, volume~1,
  pages 1715--1725.

\bibitem[\protect\citename{Stymne}2011]{stymne2011pre}
Sara Stymne.
\newblock 2011.
\newblock Pre-and postprocessing for statistical machine translation into
  germanic languages.
\newblock In {\em Proceedings of the ACL 2011 Student Session}, pages 12--17.
  Association for Computational Linguistics.

\bibitem[\protect\citename{Vania and Lopez}2017]{P17-1184}
Clara Vania and Adam Lopez.
\newblock 2017.
\newblock From characters to words to in between: Do we capture morphology?
\newblock In {\em Proceedings of the 2017 Annual Meeting of the ACL}, pages
  2016--2027. Association for Computational Linguistics.

\bibitem[\protect\citename{Virpioja \bgroup et al.\egroup
  }2007]{virpioja2007morphology}
Sami Virpioja, Jaakko~J V{\"a}yrynen, Mathias Creutz, and Markus Sadeniemi.
\newblock 2007.
\newblock Morphology-aware statistical machine translation based on morphs
  induced in an unsupervised manner.
\newblock {\em MT Summit XI}, 2007:491--498.

\bibitem[\protect\citename{Virpioja \bgroup et al.\egroup
  }2013]{virpioja2013morfessor}
Sami Virpioja, Peter Smit, Stig-Arne Gr{\"o}nroos, Mikko Kurimo, et~al.
\newblock 2013.
\newblock Morfessor 2.0: Python implementation and extensions for morfessor
  baseline.

\end{thebibliography}

\end{document}